\documentclass[letterpaper]{article} 
\usepackage{aaai2026}  
\usepackage{times}  
\usepackage{helvet}  
\usepackage{courier}  
\usepackage[hyphens]{url}  
\usepackage{graphicx} 
\usepackage{natbib}  
\usepackage{caption} 
\usepackage{algorithm}
\usepackage{algorithmic}

\usepackage{newfloat}
\usepackage{listings}
\DeclareCaptionStyle{ruled}{labelfont=normalfont,labelsep=colon,strut=off} 
\floatstyle{ruled}
\newfloat{listing}{tb}{lst}{}
\floatname{listing}{Listing}
\usepackage{url}            
\usepackage{booktabs}       
\usepackage{amsfonts}       
\usepackage{nicefrac}       
\usepackage{xcolor}         
\usepackage{amsmath}
\usepackage{enumitem}
\usepackage{makecell} 
\usepackage{algorithm}
\usepackage{subcaption}

\title{Efficient Post-Training Refinement of \\ Latent Reasoning in Large Language Models}
\author {
    Xinyuan Wang\textsuperscript{\rm 1},
    Dongjie Wang\textsuperscript{\rm 2},
    Wangyang Ying\textsuperscript{\rm 1},
    Haoyue Bai\textsuperscript{\rm 1}, \\ 
    Nanxu Gong\textsuperscript{\rm 1},
    Sixun Dong\textsuperscript{\rm 1},
    Kunpeng Liu\textsuperscript{\rm 3},
    Yanjie Fu\textsuperscript{\rm 1}\thanks{Corresponding author.}
}
\affiliations {
    \textsuperscript{\rm 1}Arizona State University, Tempe, USA\\ 
    \textsuperscript{\rm 2}University of Kansas, Lawrence, USA\\ 
    \textsuperscript{\rm 3}Clemson University, Clemson, USA\\
    \{xwang735, wying4, haoyueba, ngong6, sdong46, yanjie.fu\}@asu.edu, wangdongjie@ku.edu, kunpenl@clemson.edu
}

\usepackage{bibentry}

\begin{document}

\maketitle

\begin{abstract}
    Reasoning is a key component of language understanding in Large Language Models. 
    While Chain-of-Thought prompting enhances performance via explicit intermediate steps, it suffers from sufficient token overhead and a fixed reasoning trajectory, preventing step-wise refinement.
    Recent advances in latent reasoning address these limitations by refining internal reasoning processes directly in the model’s latent space, without producing explicit outputs.
    However, a key challenge remains: how to effectively update reasoning embeddings during post-training to guide the model toward more accurate solutions.
    To overcome this challenge, we propose a lightweight post-training framework that refines latent reasoning trajectories using two novel strategies:
    (1) Contrastive reasoning feedback, which compares reasoning embeddings against strong and weak baselines to infer effective update directions via embedding enhancement;
    (2) Residual embedding refinement, which stabilizes updates by progressively integrating current and historical gradients, enabling fast yet controlled convergence.
    Extensive experiments and case studies are conducted on five reasoning benchmarks to demonstrate the effectiveness of the proposed framework.
    Notably, a +5\% accuracy gain on MathQA without additional training. 
\end{abstract}

\begin{links}
    \link{Code}{https://github.com/anord-wang/Lateng-Reasoning}
\end{links}

\section{Introduction}

Reasoning serves as a fundamental capability in Large Language Models (LLMs), enabling them to comprehend prompts and effectively solve complex tasks.
Existing approaches, such as Chain-of-Thought (CoT)~\cite{wei2022chain} and ReAct~\cite{yao2023react}, guide models toward correct answers by explicitly generating intermediate textual reasoning steps.
While these methods have shown effectiveness, they suffer from: (1) the explicit reasoning steps cause substantial token overhead, leading to increased computational cost; (2) the reasoning trajectory becomes fixed once the template is generated, preventing step-by-step refinement during the generation process.

Recent advances have partially addressed them by converting explicit reasoning steps into latent embeddings, enabling latent reasoning in models, such as Coconut~\cite{hao2024training}.
They represent the reasoning state using the LLM’s hidden state (i.e., “continuous thought”) and recursively feed it back into the model in the latent space to enable more effective reasoning.
However, there are two critical challenges: 
(1) the reasoning trajectory in the latent space lacks explicit directional guidance, making it difficult to ensure consistent progression toward more accurate reasoning states; 
(2) the recursive embedding updates tend to be unstable, especially across multiple reasoning steps, which may compromise both robustness and accuracy. 
These challenges motivate us to explore how reasoning embeddings can be effectively and efficiently updated during post-training to guide the model toward more accurate solutions.

To this end, we draw inspiration from two complementary lines of research.
For the challenge of providing directional guidance in reasoning embedding updates, we are inspired by reinforcement learning from human feedback (RLHF)~\cite{ouyang2022training}, where learning from relative performance comparisons has demonstrated superior efficiency and effectiveness compared to relying solely on absolute supervision.
For the challenge of stabilizing recursive updates, we take inspiration from the success of momentum-based optimization techniques in deep learning~\cite{qian1999momentum}, which demonstrate the importance of adaptively integrating historical and current information to achieve smoother and more stable convergence.

Thus, we propose a lightweight post-training framework to refine the intermediate latent reasoning embeddings, built upon two novel strategies:
\begin{itemize} [nosep]
    \item \textbf{Contrastive Reasoning Feedback Search.}  
    To infer updated directions in the latent reasoning space, we pass the current reasoning embedding through both a strong and a weak model to obtain enhanced embeddings. 
    We derive a contrastive direction by comparing the outputs of strong and weak models, and use its gradient concerning the current embedding to guide the embedding update.
    The strong model is only better relative to the weak one, and does not limit the performance of the final method.

    \item \textbf{Residual Embedding Refinement.}
    To ensure stable updates in the latent space, we blend the current reasoning embedding with its previous state using a residual weighting strategy. This interpolation smooths the transition between steps and prevents abrupt shifts in the reasoning process. As a result, the model achieves more consistent convergence across multi-step latent reasoning.
\end{itemize}

We evaluate our method through comprehensive experiments and case studies, highlighting its effectiveness, efficiency, and scalability across diverse settings. These experiments demonstrate that our strategies significantly enhance reasoning performance compared to latent-only and explicit token-based reasoning baselines. Notably, on the MathQA task, our approach improves accuracy by over 5\% compared to the original latent reasoning method. We further conduct case studies to illustrate how the latent embedding evolves step by step. The results show that the embedding progresses toward more accurate reasoning solutions.

 These empirical findings not only validate the effectiveness of our approach but also highlight its practical value.
 Our framework offers three key advantages: \textbf{Efficiency and Cost-Effectiveness.} The proposed method enhances reasoning performance via a lightweight post-training refinement process. It does not require any modification to the model architecture or parameters, enabling consistent improvements with minimal cost. \textbf{Dynamic Post-Training Adaptation.} Both components operate after training to refine the reasoning process. By preserving informative latent states and exploring better latent representations, the model dynamically adjusts its internal reasoning trajectories without requiring additional training. \textbf{Training-Free Deployment.} Our refinement procedure is entirely training-free: it relies solely on forward computation in the latent space and avoids any backpropagation or parameter updates. This makes the method easy to integrate into existing models as a plug-and-play component at the post-training stage.

\section{Preliminary}

\subsection{Problem Definition} 
We focus on complex reasoning tasks where a large language model (LLM) aims to generate a correct answer $y$ from an input question $x$, such as in math word problems, multi-hop question answering, and commonsense reasoning. 
These tasks typically require multiple inference steps, even if such steps are not explicitly annotated.
Formally, the objective is to learn a function $f: x \rightarrow y$, where intermediate cognitive states are latent and only the final answer is observed.
While optional intermediate steps can be included during training, they are often unavailable during inference.
Building on the latent reasoning framework, we represent each reasoning step as an embedding in a latent space. 
Our key contribution is to model how to efficiently explore transitions within the reasoning embedding space that lead to accurate final answers.
By capturing these latent trajectories, our approach enables the model to reason more effectively, even without explicit supervision over intermediate steps.

\subsection{Chain-of-Thought Reasoning and Its Limitations}

Chain-of-Thought (CoT) prompting~\cite{wei2022chain} and its variants~\cite{wang2022self, yao2023react} decompose reasoning into a sequence of intermediate text steps, improving model accuracy and interpretability on complex tasks.
However, CoT remains inherently limited:

\begin{itemize} [nosep]
    \item \textbf{Token-level serialization}: All reasoning steps are expressed via natural language, leading to low-dimensional, rigid representations.
    \item \textbf{Static trajectory}: Once the Chain-of-Thought prompt is fixed, the reasoning path is deterministic, with limited room for correction.
    \item \textbf{Lack of feedback}: CoT does not support internal error detection or trajectory revision unless multiple sampled paths are compared externally.
\end{itemize}

These limitations motivate our shift toward latent space reasoning. Rather than generating explicit token sequences at each step, we allow the model to evolve its internal embedding over multiple steps. This latent evolution enables the model to retain richer intermediate information, search for better latent embeddings, and adapt its reasoning process more flexibly, particularly under limited model capacity.

\section{Method}

\begin{figure}[htbp]
  \centering
  \includegraphics[width=\linewidth]{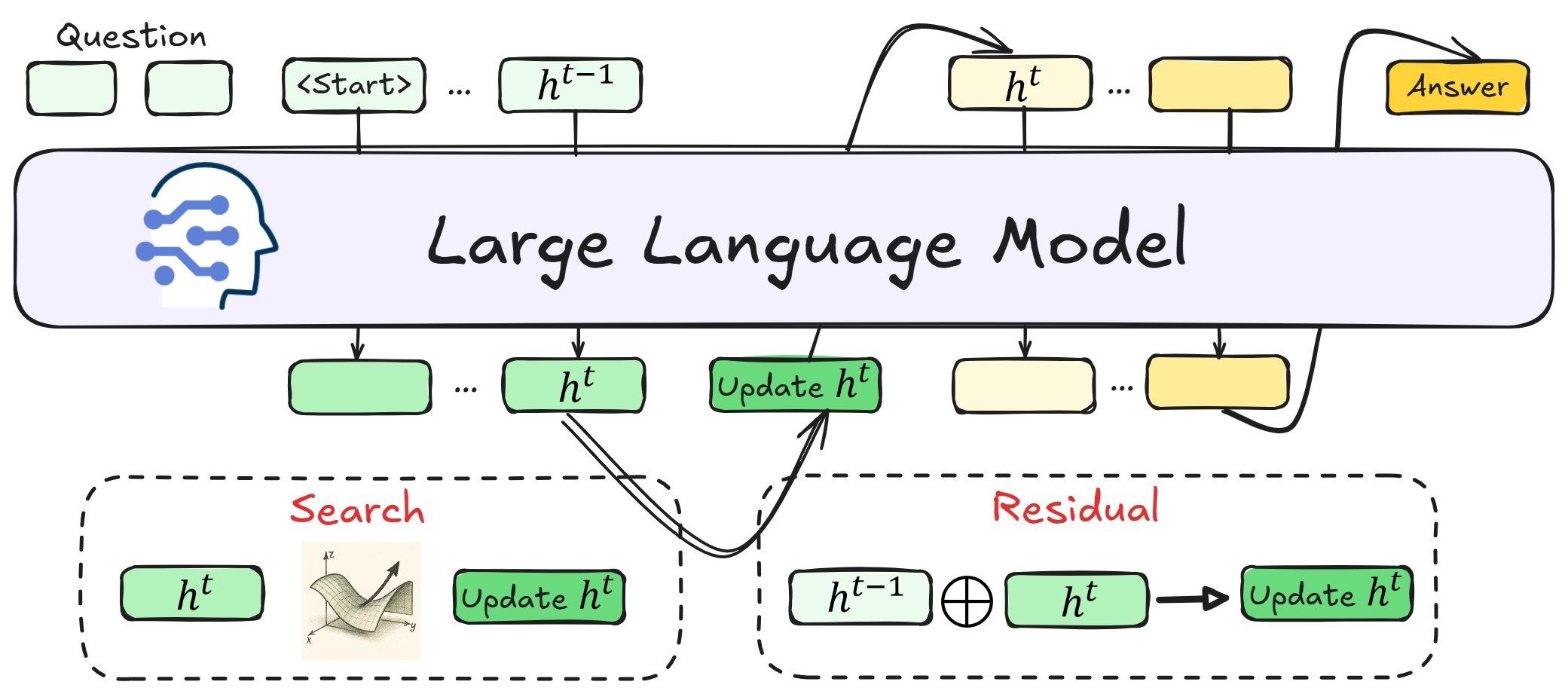}
  \caption{Overview of our reasoning framework.
   The bottom path shows how contrastive feedback first identifies the update direction for the reasoning embedding, which is then integrated through residual refinement to produce the current reasoning state.
  }
  \label{fig:framework}
\end{figure}

\textbf{Figure~\ref{fig:framework}} shows the overview of our framework. 
We introduce a general post-training latent refinement framework that enhances latent reasoning models such as Coconut~\cite{hao2024training}. Our goal is to improve reasoning stability and accuracy by augmenting the latent reasoning process with lightweight, training-free components.

In Coconut, reasoning is performed entirely in latent space without generating token-level intermediate steps, enabling compact and efficient inference. 
However, the model conducts latent updates in a fixed, feedforward manner. This end-to-end process and training cannot revise its reasoning path or retain contextual memory across steps during inference time, which limits its adaptability when errors occur. 
Adding such a correction capability during training would require large amounts of data and extensive optimization, making it expensive and less practical.

To address these problems, we propose a two-part refinement strategy applied in the post-training stage:

\begin{itemize}[nosep]
    \item \textbf{Contrastive Reasoning Feedback Search:}  This module compares reasoning outputs from a weak and a strong model to identify a contrastive improvement direction in latent space. This direction shows how the current reasoning should evolve toward stronger inference.

    \item \textbf{Residual Embedding Refinement:} This module integrates the contrastive feedback into the current reasoning using gated residual updates.
    By fusing prior context with the new signal, it preserves useful information and mitigates semantic drift across steps.

\end{itemize}

This strategy is lightweight and training-free. Both modules operate entirely in latent space using forward passes only, without any gradient updates or parameter changes. 
It allows Coconut to adjust its reasoning during inference by following a contrastive direction, while the residual update prevents overly large changes, helping to retain useful information. 
Applied at inference time, these components enable more accurate and stable reasoning with minimal computational overhead.

\subsection{Latent Reasoning Backbone}
We consider a latent reasoning framework in which the model conducts multi-step inference entirely in latent space, without producing token-level intermediate steps. 
As shown in the top path of \textbf{Figure~\ref{fig:framework}}, the input question \( x \in \mathbb{R}^L \) is first encoded into a latent embedding \( h^0 \), which  is then iteratively updated for $T$ steps using a fixed model block $f$:
\begin{equation}
h^t = f(h^{t-1}),
\end{equation}
where \( h^t \in \mathbb{R}^{L \times d} \) is the latent embedding at step \( t \).
After $T$ steps, the final state \( h^T \) is passed to a decoding head to generate the final answer. This latent-only formulation reduces token overhead and improves inference efficiency, making it particularly well-suited for small or resource-constrained models.
This backbone structure forms the basis of our reasoning framework. In our implementation, we adopt Coconut~\cite{hao2024training} as the underlying latent reasoning model, where the decoder block $f$ is derived from a pre-trained language model and kept fixed during inference.

While this setup enables compact and efficient reasoning, it still faces two key challenges:

\begin{enumerate} [nosep]
    \item \textbf{Error correction}: There is no mechanism to guide the model back when reasoning diverges, especially in the high-dimensional latent space with many possible paths.
    \item \textbf{Trajectory stability}: In multi-step reasoning, the latent trajectory may drift without memory-preserving connections, leading to unstable or inconsistent reasoning.
\end{enumerate}

To address these challenges, we introduce two lightweight post-training refinement modules, contrastive latent feedback and residual embedding refinement, that operate entirely in latent space and enhance reasoning stability and correctness without any additional training.

\subsection{Post-Training Latent Reasoning Refinement}

The latent reasoning process described in Coconut produces a series of hidden states \( h^1, h^2, ..., h^T \) by iteratively applying the model function \( f \) to an initial latent \( h^0 \). This procedure is efficient and does not generate text during reasoning. However, it performs fixed forward updates at each step and cannot revise or stabilize the reasoning trajectory.

We introduce two training-free modules that operate at the post-training stage: residual refinement and contrastive latent search. As shown in \textbf{Figure~\ref{fig:framework}}, these components are applied during inference and operate on each latent state. Both modules build on the latent-only structure of Coconut and require no access to intermediate tokens. They apply directly to the latent embeddings produced in each step.

Each step starts from the output of the Coconut-style update \( h^t = f(h^{t-1}) \), and applies a residual preservation update followed by a contrastive adjustment. These steps are designed to stabilize and correct the latent trajectory based on human-inspired memory and comparison mechanisms.

We summarize the full inference procedure in \textbf{Algorithm~\ref{alg:latent_reasoning}} in the \textbf{Appendix~\ref{sec:appendix_lgorithm}}.

\subsubsection{Contrastive Reasoning Feedback Search}

The reasoning process can still go off track even with stable latent updates due to initial uncertainty or limited model capacity. To make the system more robust, we introduce a contrastive search mechanism that enables the model to correct its latent state without any parameter updates.

\begin{figure}[htbp]
  \centering
  \includegraphics[width=1.0\linewidth]{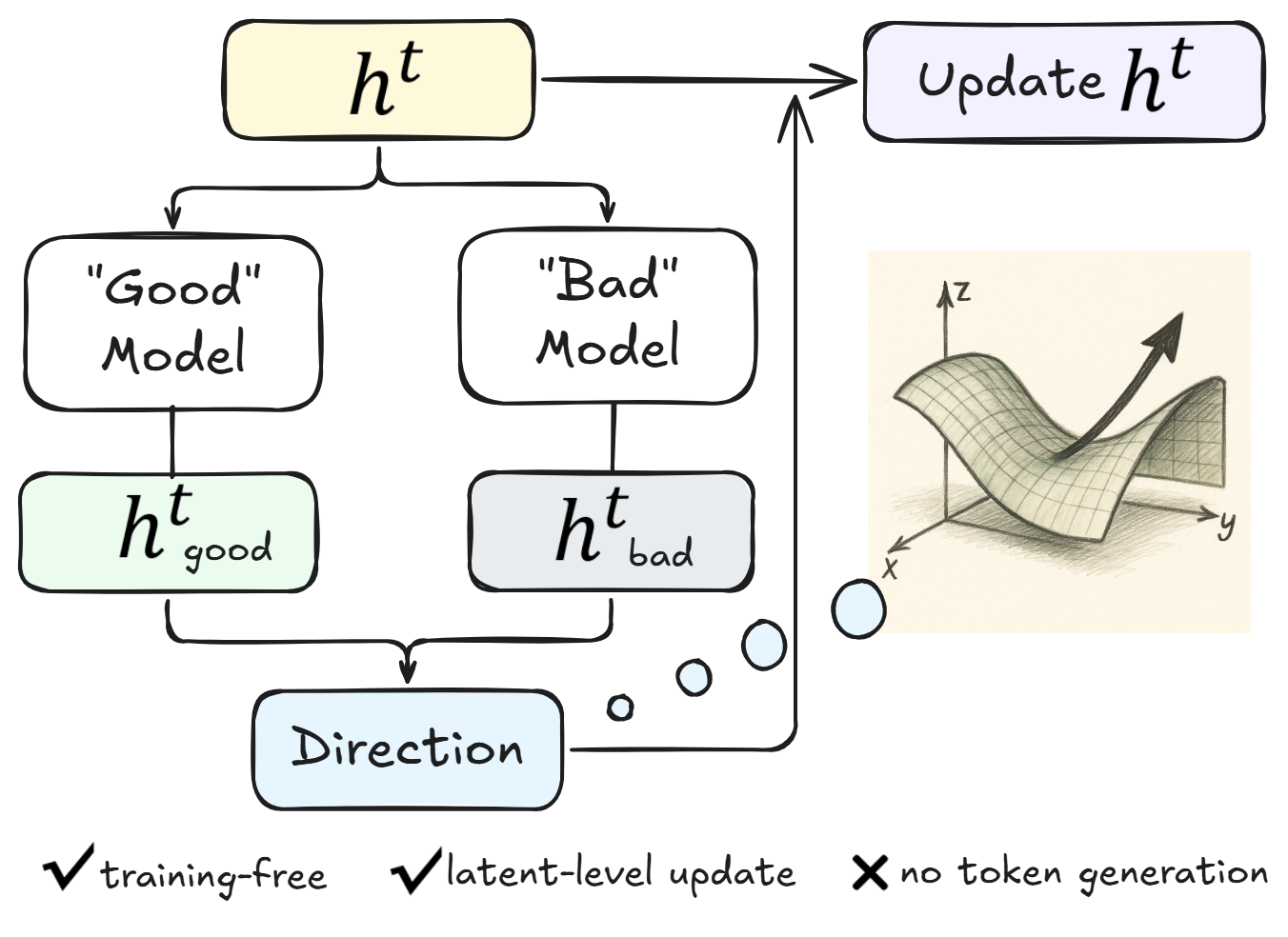}
  \caption{Contrastive Reasoning Feedback Search. We compare two models with various reasoning abilities: one stronger (``good'') and one weaker (``bad''). The direction from bad to good indicates the path to move.}
  \label{fig:search}
\end{figure}

As shown in \textbf{Figure~\ref{fig:search}}, we compare the outputs from two models of different quality at each reasoning step \( t \): a weaker model (``bad'' model, such as an early checkpoint) producing latent output \( h^t_{\text{bad}} \), and a stronger model (``good'' model, such as a later checkpoint) producing output \( h^t_{\text{good}} \).

These two models are not the final model used for latent reasoning inference (e.g., Coconut), but are intermediate checkpoints saved during the training of a standard CoT model. The distinction between ``good'' and ``bad'' is relative and only exists for the purpose of creating a directional signal in latent space. The goal is to identify a better direction to refine the current reasoning state, not to imitate or match the stronger model’s behavior.

Importantly, the ``good'' model used in this comparison is not stronger than our final model. In fact, both ``good'' and ``bad'' models are weaker than Coconut. Their role is solely to help define an updated direction based on contrast, rather than to serve as a performance reference or target.

These outputs are generated from the same input latent \( h^t \). The current latent embedding $h^t$ is then updated in a direction that reduces the distance to the good model and increases the distance from the bad model. This gives the gradient signal. Then the updated latent embedding is obtained by adjusting along the contrastive direction:

\begin{equation}
\begin{aligned}
h^t_{\text{updated}} = h^t 
+ \eta \cdot \nabla_{h^t} \big[ 
& \text{MSE}(h^t_{\text{good}}, h^t) \\
& - \text{MSE}(h^t_{\text{bad}}, h^t) 
\big].
\end{aligned}
\label{eq:search}
\end{equation}

Here, \( \text{MSE}(\cdot) \) denotes mean squared error between embeddings, and \( \eta \) is a fixed step size. 
This update is done through forward passes and gradient computation at the embedding level only. No model parameters are changed. The adjustment is lightweight and compatible with training-free inference, and can be applied once or iteratively depending on the step length.

This contrastive search provides a self-correction mechanism during reasoning. It helps the model adjust its latent trajectory without relying on external feedback or additional samples. This mimics the role of conflict monitoring and ACC adjustment in human reasoning~\cite{botvinick2001conflict}.

Currently, Coconut does not have any correction mechanism. If the reasoning goes off track, it cannot adjust or recover. CoT~\cite{wei2022chain} and Tree-of-Thoughts~\cite{yao2023tree} use external sampling or search to fix errors, but they rely on generating intermediate text. In contrast, our method updates the latent state directly using internal feedback from contrastive direction. 
This is efficient and flexible, without relying on token-level outputs or any training.

\subsubsection{Residual Embedding Refinement}

To make the latent embeddings evolve stably, we add a residual mechanism when updating the embeddings.
At each step, we update the latent state by preserving useful information from the previous step. Instead of directly replacing the latent with the new output \( f(h^{t-1}) \), we blend it with the previous state \( h^{t-1} \) using a fixed-weight residual connection:

\begin{equation}
\label{eq:residual}
h^t = \alpha \cdot h^{t-1} + (1 - \alpha) \cdot f(h^{t-1}), \quad \alpha \in [0,1]
\end{equation}
where \( \alpha \) is the memory rate. It controls how much of the previous state is kept. We use a fixed value and do not train this parameter due to the train-free setting.
This design is inspired by residual networks~\cite{he2016deep} and resembles the working memory mechanism of the human brain~\cite{koechlin2003architecture}. It allows the model to accumulate reasoning context over steps and prevents semantic drift. Without this refinement, the latent state may lose important early signals, which leads to an unstable reasoning trajectory.

In training-free settings, we cannot rely on backward gradients to correct mistakes, so it becomes more important to preserve helpful signals across steps.
The residual update plays this role by softly carrying forward past reasoning information, making the trajectory more stable.
While the contrastive reasoning search explores better directions in latent space, the residual refinement helps maintain stability during multi-step updates.

Compared to Coconut, which discards all prior hidden states at each step, our refinement preserves and integrates context, leading to more consistent and accurate reasoning.

\section{Experimental Results}
We conduct experiments to evaluate whether our latent reasoning framework improves reasoning accuracy, supports training-free deployment, and operates efficiently across diverse tasks and models. Specifically, our experiments are structured to answer the following key questions:

\noindent\textbf{Q1:}  Can our method improve reasoning performance with minimal cost, using only training-free post-processing?

\noindent\textbf{Q2:} Compared to fully retraining, can our post-training refinement achieve improvements with lower resource usage?

\noindent\textbf{Q3:}  How important are the two components—residual refinement and contrastive latent search—in contributing to performance improvement?

\noindent\textbf{Q4:} Can the latent update mechanism consistently steer the model toward more accurate predictions in specific reasoning instances?

\noindent\textbf{Q5:} How robust is our method to changes in hyperparameters like memory update rate and latent search step length?

\noindent\textbf{Q6:} Can our framework generalize well across diverse language model architectures and parameter sizes?

\subsection{Experimental Setup}

\noindent\textbf{Datasets.} We evaluate our method on five representative benchmarks covering math, commonsense, and multi-hop reasoning: GSM8K~\cite{cobbe2021training}, MathQA~\cite{amini2019mathqa}, AQUA-RAT~\cite{ling2017program}, StrategyQA~\cite{geva2021did}, and ProsQA~\cite{hao2024training}. Dataset descriptions are provided in \textbf{Appendix~\ref{sec:appendix_datasets}}.
\smallskip

\noindent\textbf{Models.} We use GPT-2 base (117M) to conduct the majority of experiments. We also try two other well-recolonized open-source language models: Qwen-2.5 1.5B and LLaMA-3.2 3B. 
For contrastive search, we utilize checkpoints from different training stages in the CoT method as ``good'' or ``bad'' references.
\smallskip

\noindent\textbf{Baselines.} We compare our method with the following approaches: (1) \textbf{No-CoT}: directly trains GPT-2~\cite{radford2019language} to generate the final answer without any intermediate reasoning steps; (2)\textbf{Chain-of-Thought (CoT)}~\cite{wei2022chain}: standard step-by-step natural language reasoning approach; (3) \textbf{Coconut}~\cite{hao2024training}: latent reasoning without search or refinement; (4) \textbf{Ours}: latent reasoning with contrastive search and residual refinement.
\smallskip

\noindent\textbf{Evaluation.} We report exact match precision, averaged on 3 random seeds. No fine-tuning is used; our method improves reasoning only through forward latent-space updates. Full implementation details are provided in \textbf{Appendix~\ref{sec:appendix_hyperparams}}.

\subsection{Q1: Overall Reasoning Performance}

\begin{figure}[th]
  \centering
  \includegraphics[width=\linewidth]{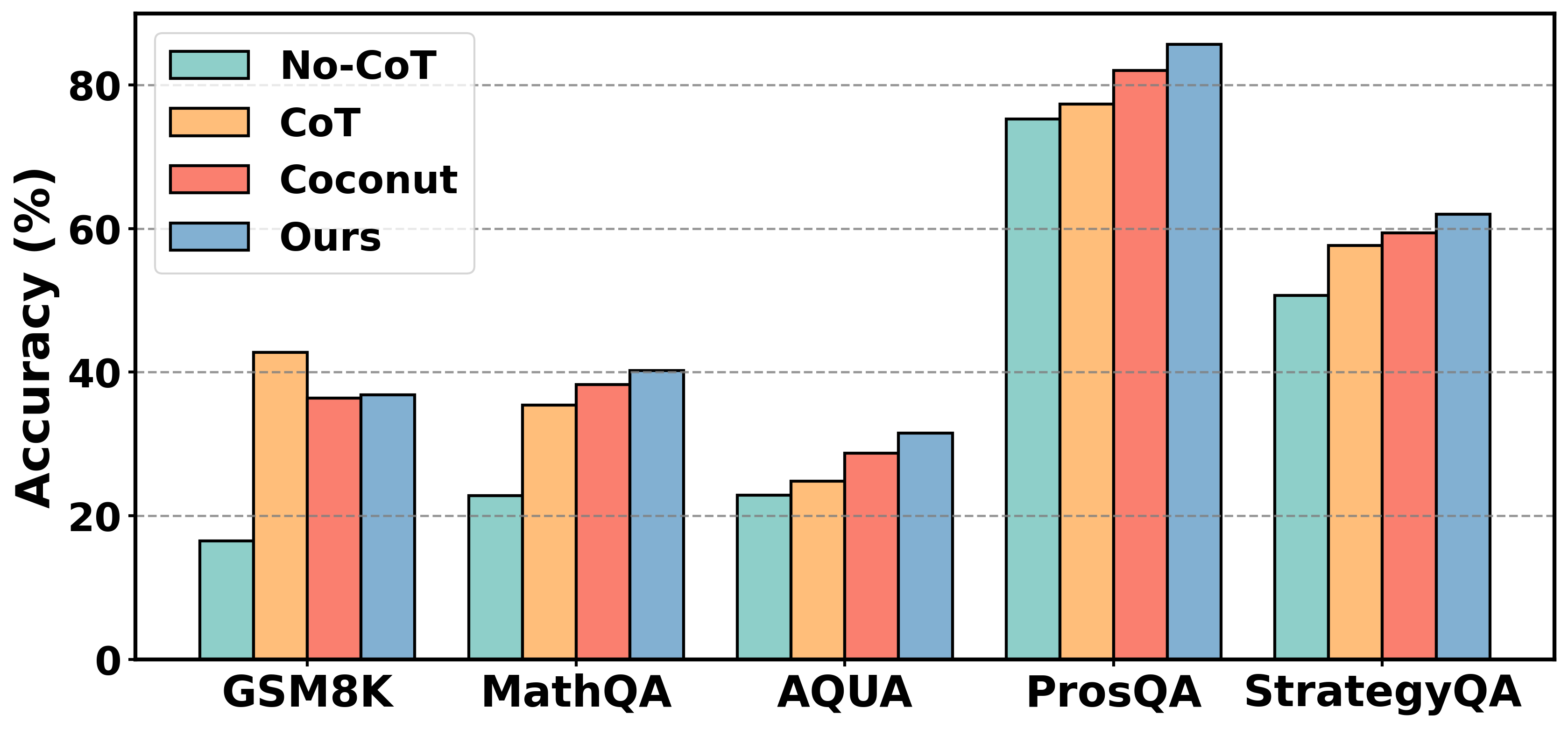}
  \caption{Accuracy (\%) of different reasoning methods across five benchmarks.}
  \label{fig:main_results}
\end{figure}

To answer Q1,  we compare our method with the baseline algorithms on five benchmarks to test whether our latent-space refinement strategy improves accuracy with small costs (train-free). \textbf{Figure~\ref{fig:main_results}} shows the reasoning accuracy (Y-axis) of our method on five benchmark tasks (X-axis).

\textbf{Figure~\ref{fig:main_results}} shows that our method consistently outperforms latent-only reasoning (Coconut) and explicit token-based reasoning (CoT) on four out of five tasks.
Notably, on the tasks of MathQA and AQUA—both that involves multi-step numerical and symbolic reasoning, our model achieves +1.95\% and +2.76\% absolute gains (not relative improvement ratio) over Coconut in terms of reasoning accuracy, respectively. On ProsQA and StrategyQA, which focus more on structured logical reasoning and commonsense composition, we observe absolute gains of +3.67\% and +2.63\% in reasoning accuracy.

The results highlight that: 
(1) Although CoT produces explicit thought steps, it suffers from verbosity and error accumulation, especially in non-math tasks. 
(2) Coconut improves reasoning by operating in a compact latent space, but lacks error correction and stable refinement. 
(3) Our method combines both advantages through residual preservation and feedback-driven search, thus resulting into more reliable and generalizable reasoning, especially in settings where intermediate steps are implicit or hard to verbalize.

One exception is GSM8K, where CoT remains the most effective method (42.76\%). This is because GSM8K contains complex arithmetic problems that require symbolic calculations. Humans rely on written steps for such tasks rather than purely mental computation. Without external tools or explicit formulas, latent reasoning struggles to handle long-chain numerical operations. 
In contrast, MathQA, although also math-focused, has more structured and templated problems and is multiple-choice, which makes the task easier compared to GSM8K's open-ended answers. 
Under such settings, our method can benefit more from embedding refinement and soft memory tracking. 
This highlights the differences in cognitive demands between math datasets and suggests that combining latent reasoning with symbolic or tool-augmented components may be a future direction.

\subsection{Q2: Training vs. Inference Performance}
To answer Q2, we compare our train-free method with other training strategies to examine the accuracy of reasoning and the usage of resources in the ProsQA dataset. 
We train a base Coconut model for 30 epochs, then evaluate four settings: (1) using the original Coconut model, (2) only applying the latent reasoning during inference, (3) continuing training for 10 more epochs with our latent reasoning, and (4) applying our latent reasoning during both training and inference.

\begin{figure}[t]
  \centering
  \includegraphics[width=\linewidth]{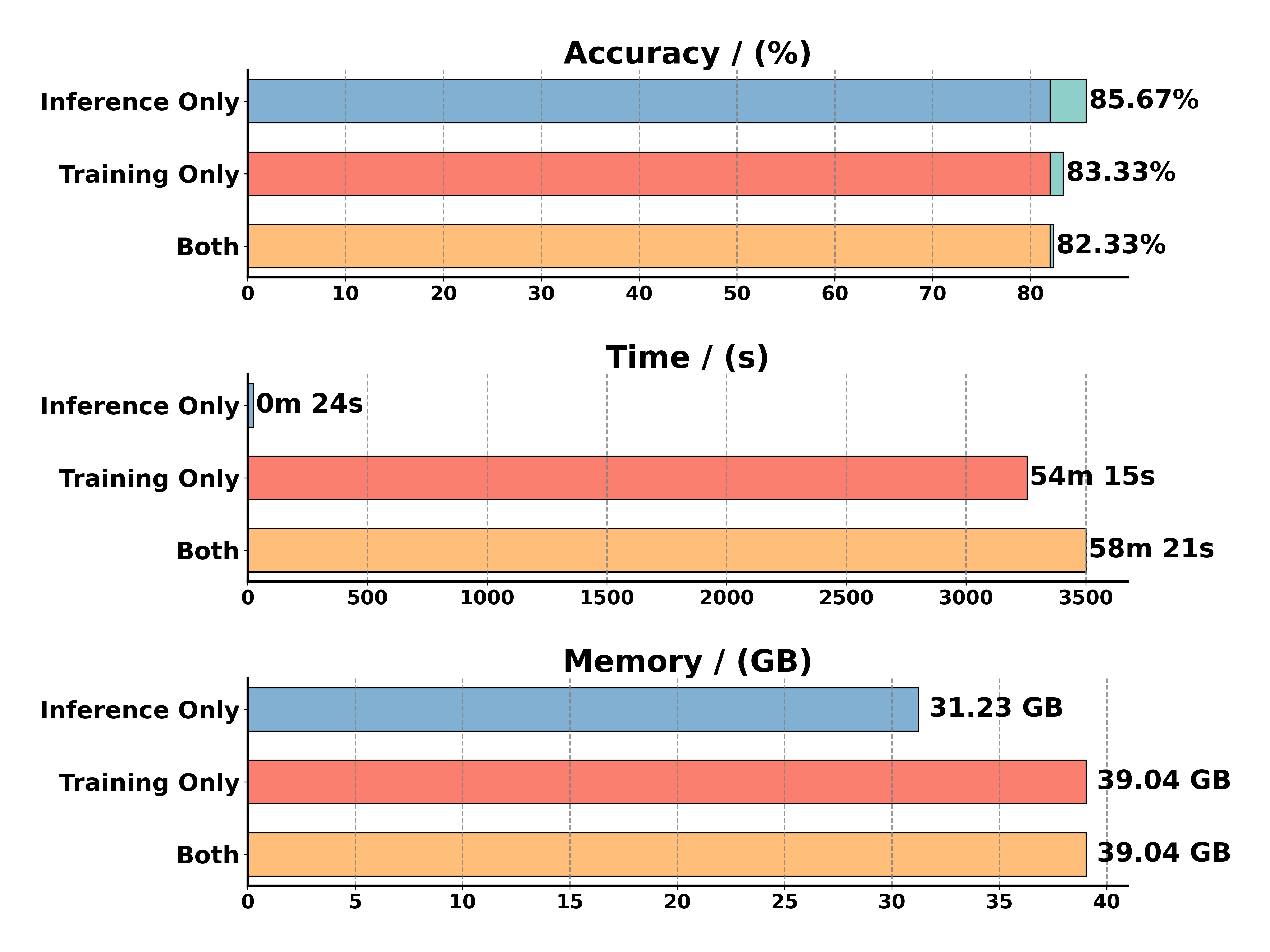}
  \caption{Latent Reasoning in Training vs. Inference.}
  \label{fig:train_infer}
\end{figure}

\textbf{Figure~\ref{fig:train_infer}} shows that our method achieves the best accuracy (+4.47\%) when applied only during inference, with minimal overhead (24 seconds, 31.23GB memory). 
In contrast, continuing training with latent reasoning adds significant time (54+ minutes) and memory cost (39.04GB), yet leads to smaller accuracy gains (+1.63\%). 

Using the method in both training and inference further increases the cost but gives almost no improvement.
This result may seem counter-intuitive at first. But we observed the same pattern on three datasets, so we believe it is consistent.
Our understanding is: during training, the refinement only affects the forward pass. It does not go into the backward gradient, so the model may not fully learn how to use it. With limited training data, this extra signal may even confuse the model or hurt convergence.
Then, during inference, if we apply refinement again, the model receives a second adjustment. This may ``overshoot'' the correct direction and make the reasoning worse. That could explain why training + inference gives smaller gains than inference-only.

These results highlight the efficiency of our strategy: without any backpropagation or weight updates, our inference-only setup improves performance while saving training time and GPU resources, and without the need for further training.

\textbf{Insights.} Our method avoids expensive re-training, requires only forward computation, and still brings accuracy improvement. 
It is lightweight and improves the reasoning trajectory without changing the model parameters. 
These computational benefits make it practical for deployments in low-resource or frozen-model scenarios.

\subsection{Q3: Study of Contrastive Search and Residual Refinement}
To answer Q3, we remove the the contrastive search and residual refinement one by one to understand the role of each technical component.

\begin{table}[tbh]
\centering
\caption{Ablation study on MathQA. We show the impact of each component in our method.}
\label{tab:ablation}
\resizebox{1.0\linewidth}{!}{
\begin{tabular}{lcc}
\toprule
\textbf{Variant} & \textbf{Accuracy (\%)} & \textbf{Gain (\%)} \\
\midrule
Latent only & 38.25 & - \\
+ Residual refinement & 40.02 & +4.63 \\
+ Latent Search & 39.79 & +4.03 \\
+ Residual + Search (ours) & \textbf{40.20} & \textbf{+5.10} \\
\bottomrule
\end{tabular}
}
\end{table}

\textbf{Table~\ref{tab:ablation}} highlights the importance of residual refinement and contrastive search in our framework. 
Compared to the Coconut baseline with only latent thoughts, incorporating residual connections yields a +4.63\% improvement in accuracy. 
This suggests that preserving and gradually refining previous latent states helps stabilize reasoning trajectories across steps.
Using only contrastive search leads to a +4.03\% gain, showing that updating the current latent state with better directions enables the model to recover from suboptimal reasoning.
Finally, integrating both residential connections and contrastive search results in the best performance (+5.10\%). 
This shows that stable memory evolution and dynamic search together form a lightweight and training-free latent reasoning mechanism that maintains contexts, detects errors, and refines internal representations without relying on explicit intermediate language tokens.

\subsection{Q4: A Step-wise Case Study}
To answer Q4, we visualize the impacts of latent reasoning refinement on answer prediction to better understand how our method improves reasoning using the MathQA dataset.

\begin{figure}[htbp]
    \centering
    \includegraphics[width=1.0\linewidth]{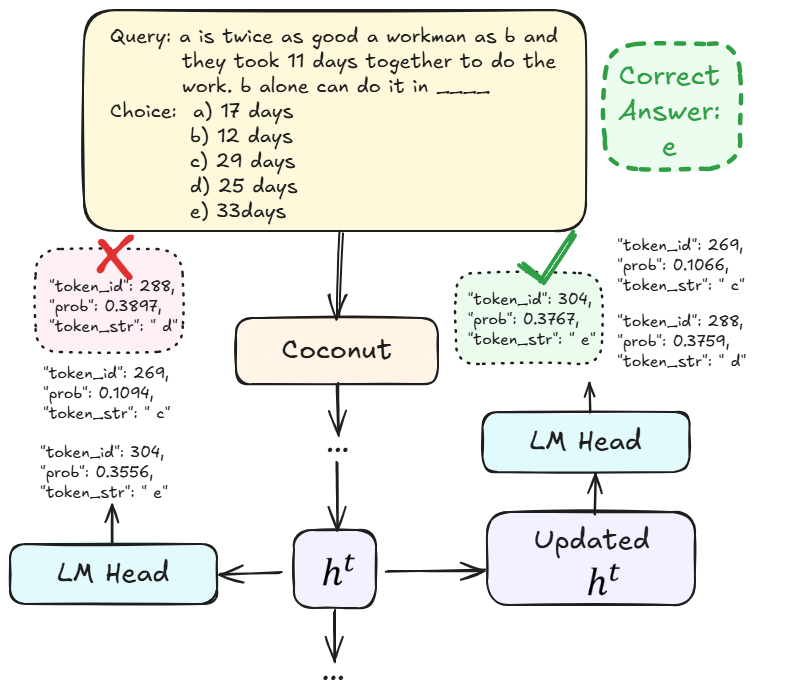}
    \caption{Case study: our method adjusts the latent embedding to reach the correct answer.}
    \label{fig:step_wise}
\end{figure}

\textbf{Figure~\ref{fig:step_wise}} shows the baseline Coconut produces a latent state $h_t$ and predicts the wrong choice (``d'') with the highest probability. 
After applying our latent refinement, the updated embedding $h_t'$ leads to the correct prediction (``e'').
One possible explanation is: contrastive search adjusts the latent state using information from reference models, while residual refinement helps to preserve internal information across steps. 
Both are used in forward steps only without additional training. 
The model dynamically adjusts its internal representation to align with the correct reasoning trajectory.

\textbf{Insights.} The case study illustrates how our method enables internal latent correction before decoding. Rather than relying on external tokens or explicit logic, the model self-adjusts in latent space, which reflects how humans reconsider their thoughts before answering. We provide three more examples in \textbf{Appendix~\ref{app:casestudy}}.

\subsection{Q5: Study of Hyperparameter Sensitivity}

To answer Q5, we examine the sensitivity of two key parameters of contrastive search and residual refinement using the MathQA and AQUA datasets: (1) \textbf{Search Step Length} ($\eta$ in \textbf{Equation~\eqref{eq:search}}): the step of each latent update in the contrastive search; (2) \textbf{Memory Rate} ($\alpha$ in \textbf{Equation~\eqref{eq:residual}}): how much previous latent state is memorized across reasoning steps.
For each query, we perform 3 rounds of latent refinement, each of which includes one residual update and one search update, and fix other factors for fair comparisons.

\begin{figure}[htbp]
    \centering
    \begin{subfigure}[t]{0.48\linewidth}
        \centering
        \includegraphics[width=\linewidth]{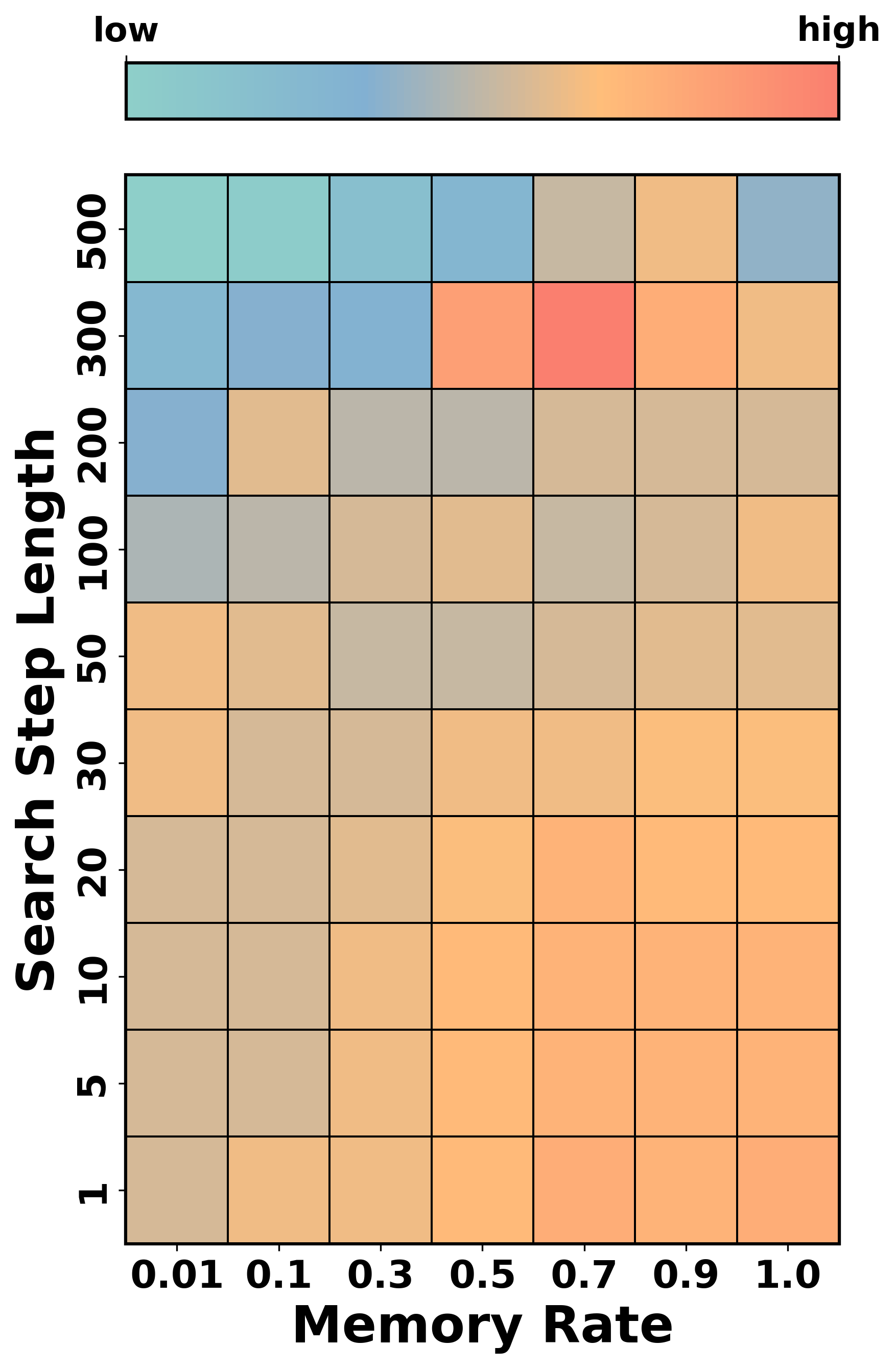}
        \caption{Sensitivity on MathQA}
        \label{fig:mathqa_sensitivity}
    \end{subfigure}
    \hfill
    \begin{subfigure}[t]{0.48\linewidth}
        \centering
        \includegraphics[width=\linewidth]{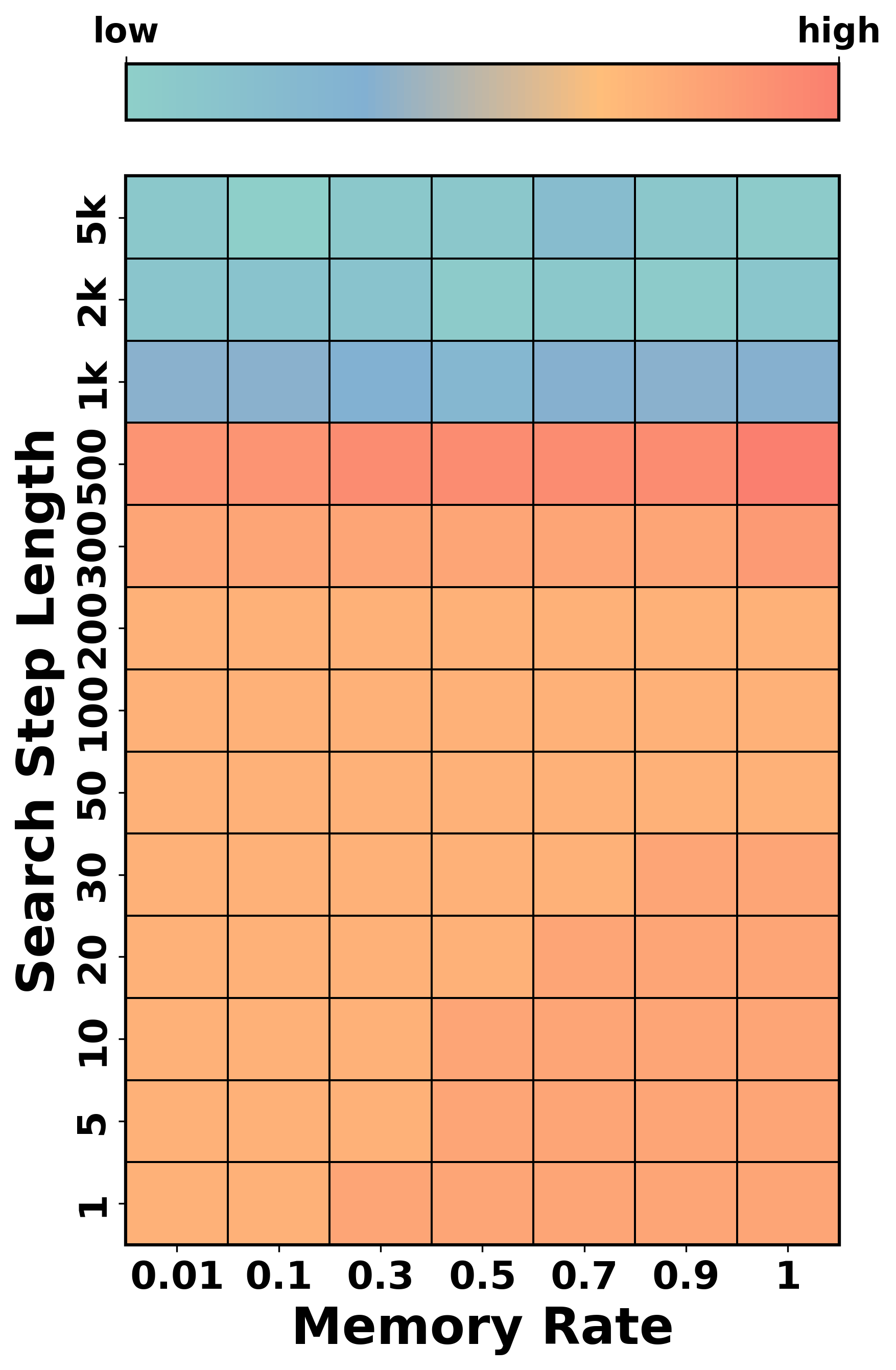}
        \caption{Sensitivity on AQUA}
        \label{fig:aqua_sensitivity}
    \end{subfigure}
    \caption{Hyperparameter sensitivity on two datasets. The colors measure reasoning accuracy.}
    \label{fig:parameter_sensitivity}
\end{figure}

\textbf{Figure~\ref{fig:mathqa_sensitivity}} shows the heat map of accuracy in MathQA on memory rates and search steps, where red indicates high accuracy and green indicates low accuracy.
Overall, the reasoning accuracy increases when the memory rate increases, but is less affected by the search step length.
This suggests that solving mathematical problems benefits more from stable accumulation of prior steps and residual refinement (i.e., memory rate). A stronger memory allows the model to preserve intermediate reasoning steps for better reasoning.
\textbf{Figure~\ref{fig:aqua_sensitivity}} shows the opposite trend. AQUA is less sensitive overall, but is more affected by the search step size. This suggests that common-sense QA relies more on adaptive correction than long-term memory, so flexible updates in latent space have a greater impact than memory accumulation.

\textbf{Insights.} The results show that different reasoning tasks may rely on different cognitive mechanisms. Math-heavy tasks need better memory. QA tasks benefit more from flexible error correction. Our framework allows both without extra training.

\subsection{Q6: Generalization Across Backbones}

To answer Q6, we evaluate the generalization of our framework on different LLM backbones across architectures and sizes: GPT-2 base (117M), Qwen-2.5 1.5B, and LLaMA-3.2 1B models.

\textbf{Table~\ref{tab:backbone}} shows that our method improves accuracy over the baseline on all three backbones. For example, on Qwen and LLaMA, the base accuracy is 42.29\% and 39.30\%, and our method improves it to 43.04\% and 41.10\%, with gains of +1.77\% and +4.58\%, respectively. This confirms that our latent refinement strategy can work across different model architectures and sizes.

The method is simple and general. It does not require any changes to the model architecture and applies equally to small and medium models. It also adds little overhead. Training time and memory increase with model size as expected, but inference remains efficient — all models complete in under 7 minutes and under 24 GB of memory.

Our approach is especially practical for frozen or resource-limited setups. It can be used on top of any existing language model to improve reasoning, without re-training or modifying model weights.

\begin{table}[h]
\centering
\caption{Comparison across LLMs on MathQA.}
\label{tab:backbone}
\resizebox{\columnwidth}{!}{
\begin{tabular}{lccc}
\toprule
& \textbf{GPT-2} & \textbf{Qwen} & \textbf{LLaMA} \\
\midrule
Parameters       & 117M       & 1.5B     & 3B \\
Accuracy (\%)    & 40.20      & 43.04       & 41.10 \\
Latent Only Acc (\%)    & 38.25      & 42.29       & 39.30 \\
Gain (\%)    & +5.10      & +1.77       & +4.58 \\
Train Time (min) & 18.95      & 95.80       & 78.38 \\
Train Mem (GB)   & 45.68      & 45.33 (LoRA)        & 45.74 (LoRA) \\
Infer Time (min) & 4.27       & 6.31       & 5.83 \\
Infer Mem (GB)   & 22.64      & 23.12       & 22.84 \\
\bottomrule
\end{tabular}
}
\end{table}

\subsection{Study of Token Efficiency}
Latent reasoning avoids generating intermediate natural language steps (i.e., thoughts), thus, reduce output token number. 
We compare the average number of generated tokens between Chain-of-Thought (CoT) prompting and our latent reasoning method on the MathQA and AQUA datasets.

\begin{table}[hb]
\centering
\caption{Average generated tokens per query.}
\label{tab:token_saving}
\resizebox{0.8\columnwidth}{!}{
\begin{tabular}{lccc}
\toprule
& \textbf{CoT} & \textbf{Latent} & \textbf{Reduction} \\
\midrule
MathQA & 66.71 & 5.02 & 92.47\% \\
AQUA & 72.73 & 5.31 & 92.65\% \\
\bottomrule
\end{tabular}
}
\end{table}

\textbf{Table~\ref{tab:token_saving}} shows that latent method reduces token usage by over 92\% on both datasets. This is because latent steps are performed in a latent space and don't produce textual outputs at each reasoning step.
These results suggest that latent reasoning is not only more compact but also more cost-efficient during inference. This makes it particularly suitable for deployment in resource-constrained settings or large-scale usage.

\section{Related Work}

\subsection{Chain of Thought Reasoning}

Chain-of-Thought (CoT) prompting enhances reasoning in LLMs by decomposing complex problems into step-by-step textual reasoning traces~\cite{wei2022chain,kojima2022large}.  
Later works improve CoT with better aggregation (e.g., self-consistency~\cite{wang2022self}), scalable prompt generation~\cite{zhang2022automatic}, and tree-structured exploration like Tree-of-Thoughts~\cite{yao2023tree}.  
Prompting strategies have also been optimized through complexity-aware design~\cite{fu2022complexity} and iterative demonstration~\cite{nye2021show}.  
These techniques have been widely applied to domains such as math~\cite{zhou2022least}, commonsense~\cite{huang2022language}, and symbolic logic~\cite{liu2023glore}, showing broad improvements.  
Extensions such as Selection-Inference~\cite{creswell2022selection}, Program-of-Thoughts~\cite{chen2022program}, and Multimodal-CoT~\cite{lu2022learn} further combine CoT with selection mechanisms, symbolic programs, or visual reasoning inputs.

\subsection{Latent-Space Reasoning in LLMs}

Recent work proposes moving reasoning from token-level generation to latent space updates, enabling more compact and abstract reasoning.  
Coconut~\cite{hao2024training} is a foundational method that replaces intermediate token outputs with internal latent states that evolve step by step.  
Other approaches expand this idea through latent sampling (LaTRO~\cite{chen2024language}), self-training to uncover latent reasoning (SERT~\cite{zhang2025self}), or expectation-maximization loops for iterative reasoning~\cite{ruan2025reasoning}.  
Looped transformers~\cite{saunshi2025reasoning} reuse a smaller model multiple times to simulate deep reasoning trajectories.  
In applied domains, ReaRec~\cite{tang2025think} introduces latent multi-step reasoning into recommender systems, demonstrating the broader utility of latent-state methods beyond language tasks.

\section{Conclusion Remarks}

We present a latent reasoning framework for LLM by introducing residual refinement and contrastive latent search to improve stability and enable dynamic self-correction without retraining. 
The new framework operates in a training-free, plug-and-play manner. 
Experiments show that our method improves accuracy by 2–5\% over latent-only reasoning across five benchmarks, with up to 7.7\% gain on ProsQA. 
The improvement demonstrates the effectiveness of latent refinement with internal information.

Our method is simple, efficient, and general. It works across different model architectures and sizes, and adds minimal overhead during inference. 
Even without generating intermediate text or using extra supervision, LLMs can still adjust their internal states through small updates in the latent space. 
This shows a new way to make models correct themselves at the embedding level, which can be useful for frozen models or low-resource settings.

\section{Acknowledgments}
Dr. Yanjie Fu is supported by the National Science Foundation (NSF) via the grant numbers: 2426340, 2416727, 2421865, 2421803.
Dr. Kunpeng Liu is supported by the National Science Foundation (NSF) via the grant numbers 2550105, 2550106, and 2242812.

\bibliography{aaai2026}

\appendix

\section{Algorithm Overview}
\label{sec:appendix_lgorithm}

This is the pseudo-code of our train-free framework.

\begin{algorithm}[htbp]
\caption{Latent Reasoning with Residual Refinement and Contrastive Search (Inference-Time)}
\label{alg:latent_reasoning}
\begin{algorithmic}[1]
\REQUIRE Question input $x$, reasoning steps $T$, pretrained model $f$, residual weight $\alpha$, search step size $\eta$, strong model $f_{\text{good}}$, weak model $f_{\text{bad}}$
\STATE Encode input $x$ into initial latent state $h^0$
\FOR{$t = 1$ to $T$}
    \STATE Compute latent update: $\Delta h^t \leftarrow f(h^{t-1})$
    \STATE Residual refinement: $h^t \leftarrow \alpha \cdot h^{t-1} + (1 - \alpha) \cdot \Delta h^t$
    \IF{Contrastive search is enabled}
        \STATE Get strong model output: $h^t_{\text{good}} \leftarrow f_{\text{good}}(h^t)$
        \STATE Get weak model output: $h^t_{\text{bad}} \leftarrow f_{\text{bad}}(h^t)$
        \STATE Compute gradient: $g^t \leftarrow \nabla_{h^t} \left[ \text{MSE}(h^t, h^t_{\text{good}}) - \text{MSE}(h^t, h^t_{\text{bad}}) \right]$
        \STATE Contrastive update: $h^t \leftarrow h^t + \eta \cdot g^t$
    \ENDIF
\ENDFOR
\STATE Decode final latent $h^T$ into answer $y$
\RETURN Answer $y$
\end{algorithmic}
\end{algorithm}

\section{Code Repository}

Our method is based on Coconut~\cite{hao2024training}. Coconut codes are available at: \url{https://github.com/facebookresearch/coconut}.

The codes and data used in this paper are publicly available at this anonymous repository: \url{https://github.com/anord-wang/Lateng-Reasoning}.

\section{Dataset Descriptions}
\label{sec:appendix_datasets}

\noindent
\textbf{GSM8K}~\cite{cobbe2021training}\footnote{\url{https://huggingface.co/datasets/openai/gsm8k}}: A dataset of 8.5K high-quality grade school math word problems created by human problem writers. It is divided into 7.5K training and 1K test examples. Each problem requires 2–8 steps of basic arithmetic reasoning (addition, subtraction, multiplication, division) and includes a detailed natural language solution explanation.

\noindent
\textbf{MathQA}~\cite{amini2019mathqa}\footnote{\url{https://huggingface.co/datasets/allenai/math_qa}}: A dataset containing approximately 37.2K math word problems derived from AQuA. Each problem is annotated with a programmatic solution template using a domain-specific language comprising 58 operators. The dataset covers various mathematical domains, including percentages, geometry, and linear equations, and provides multiple-choice answers.

\noindent
\textbf{AQUA-RAT}~\cite{ling2017program}\footnote{\url{https://huggingface.co/datasets/deepmind/aqua_rat}}: A large-scale dataset consisting of approximately 100K algebraic word problems. Each question is accompanied by a step-by-step natural language rationale explaining the solution process. The dataset is designed to train models capable of generating both the solution and the explanatory rationale.

\noindent
\textbf{StrategyQA}~\cite{geva2021did}\footnote{\url{https://huggingface.co/datasets/voidful/StrategyQA}}: A question-answering benchmark comprising 2,780 examples. Each question requires implicit multi-hop reasoning, where the necessary reasoning steps are not explicitly stated and must be inferred. The dataset includes decompositions and evidence paragraphs for each question.

\noindent
\textbf{ProsQA}~\cite{hao2024training}: A diagnostic dataset designed to evaluate models' planning abilities. It features examples that require multi-step reasoning and the ability to handle distractors. Each question is paired with a human-written rationale, facilitating the assessment of models' reasoning processes.

\section{Chain-of-Thought Data Construction}
\label{sec:cot_data}

To compare different reasoning paradigms, we construct CoT-style training and evaluation data from datasets that provide intermediate reasoning steps. Each example contains three parts: a \textbf{question} $x$, a list of \textbf{steps} $s = \{s_1, s_2, ..., s_n\}$ describing the reasoning trace, and a final \textbf{answer} $y$.

We format the data differently depending on the method:

\begin{itemize}[nosep]
    \item \textbf{No-CoT (Direct QA)}: The model is trained to map the input question directly to the answer. Only $(x, y)$ pairs are used, without any reasoning trace.

    \item \textbf{CoT (Textual Supervision)}: The model is trained to generate both the reasoning steps and the final answer as a single sequence. The input is $x$, and the output is $\texttt{join}(s_1, ..., s_n) + y$. This helps the model learn to reason explicitly in text.

    \item \textbf{Latent CoT (Latent Reasoning)}: We follow the Coconut~\cite{hao2024training} setup. The model is trained to generate the answer $y$ from input $x$, but the reasoning steps $s$ are used as supervision in the latent space. They are not generated as tokens, but guide the intermediate hidden representations.
\end{itemize}

This unified setup allows us to study how different forms of reasoning supervision affect model performance, and how latent-space reasoning compares to explicit token-level chains.





\section{Implementation and Hyperparameters}
\label{sec:appendix_hyperparams}

\paragraph{Optimization.}  
All models are trained with the AdamW optimizer, using a learning rate of $3 \times 10^{-5}$, batch size 16, and weight decay 0.01. Training runs for 50 epochs unless specified otherwise. When training is staged (e.g., for latent reasoning), we use 30 epochs for initial training and 10 for optional refinement. Gradients are accumulated every step (no gradient accumulation).

\paragraph{Model Backbones.}  
We primarily use GPT-2 (117M) as the language model. The pretrained model is loaded from \texttt{openai-community/gpt2}. For contrastive latent search, we define a ``bad'' checkpoint (early-stage model) and a ``good'' checkpoint (later-stage model), loaded separately and used in forward-only search (no weight update).

\paragraph{Latent Reasoning Configuration.}  
All reasoning happens in the latent space:
\begin{itemize}[nosep]
    \item \textbf{Latent update steps:} 3 iterations per query.
    \item \textbf{Search step size ($\eta$):} from 1 to 5000: controls the gradient step in contrastive search.
    \item \textbf{Residual memory rate ($\alpha$):} from 0.01 to 1: controls how much information from the previous latent state is retained.
\end{itemize}

Both residual update and contrastive search are applied at each reasoning step without gradient backpropagation. The full process is training-free and runs with forward computation only.

\paragraph{Tokenization and Decoding.}  
We use the native tokenizer of each backbone model. For latent reasoning, we add special tokens such as \texttt{<|latent|>} to mark latent segments. The max token limit is set to 64 for math tasks and 128 for QA tasks.

\paragraph{Reproducibility.}  
All experiments are run with seed 0. Our setup supports distributed training and evaluation across 4 GPUs. Each configuration, model path, and dataset split is version-controlled and specified in the released code.

\section{Experimental Environment}
All experiments were conducted on the Ubuntu 22.04.3 LTS operating system, with a 13th Gen Intel(R) Core(TM) i9-13900KF CPU, 128GB of system RAM, and four NVIDIA RTX A6000 GPUs (each with 48GB of VRAM). The experiments were implemented using Python 3.11.5 and PyTorch 2.0.1.

\section{Generalization to Unseen Tasks}
\label{app:generalization}

In this section, we analyze whether latent reasoning learned from one task can generalize to another without re-training. This evaluates the portability of internal latent-space dynamics across domains.

We conducted a controlled experiment where all models were trained on GSM8K and directly tested on MathQA. Both datasets involve multi-step math word problems but differ in format, vocabulary, and complexity. For reference, we also evaluate models trained directly on MathQA.

\begin{table}[htbp]
\centering
\caption{Generalization to MathQA: Accuracy of models trained on different datasets}
\label{tab:generalization}
\resizebox{0.8\linewidth}{!}{
\begin{tabular}{lccc}
\toprule
\textbf{Model} & \makecell[c]{Trained on\\MathQA} & \makecell[c]{Trained on\\GSM8K} & \textbf{Difference} \\
\midrule
CoT      & 35.42 & 0.00 & -35.42 \\
Coconut  & 38.25 & 0.00 & -38.25 \\
Ours     & \textbf{40.20} & 0.00 & \textbf{-40.20} \\
\bottomrule
\end{tabular}
}
\end{table}

\paragraph{Observation.}  
As shown in Table~\ref{tab:generalization}, none of the models generalize successfully from GSM8K to MathQA. Accuracy drops to 0 across CoT, Coconut, and our latent reasoning framework.

\paragraph{Analysis.}  
These results highlight the challenge of cross-task generalization in small and medium-sized LLMs. Despite using latent representations and post-training refinement, our method—like CoT and Coconut—fails to transfer reasoning skills between datasets. We believe this is due to two factors: (1) models like GPT-2 lack the scale and capacity needed for emergent generalization~\cite{wei2022emergent}; (2) our method is designed to refine reasoning trajectories within a task, not to perform domain adaptation. While effective in-distribution, the framework does not substitute for task-specific training when reasoning patterns differ across domains.

\section{More Output Case Studies}
\label{app:casestudy}

We present additional examples to illustrate how our latent reasoning refinement improves model predictions. In each case, the base Coconut model produces a suboptimal answer, while our method adjusts the latent representation to arrive at the correct one. These examples further demonstrate the effectiveness of residual and contrastive updates at inference time.

\subsection{Example 1}

\begin{figure}[htbp]
    \centering
    \includegraphics[width=0.9\linewidth]{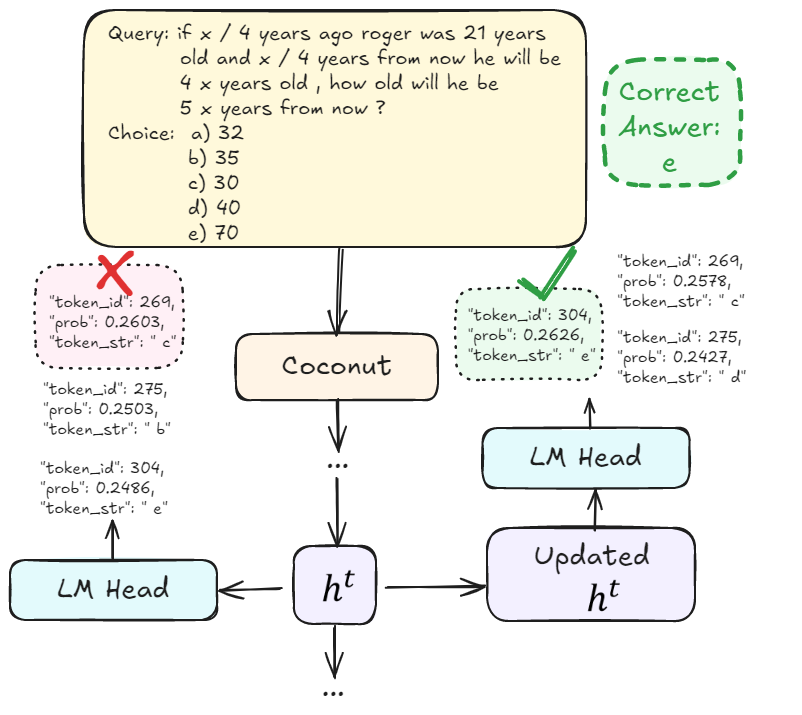}
    \caption{Additional Case study 1: our method adjusts the latent embedding to reach the correct answer.}
    \label{fig:step_case_1}
\end{figure}

In \textbf{Figure~\ref{fig:step_case_1}}, the input question requires multi-hop reasoning. The base model (Coconut) assigns the highest probability to the incorrect option ``c'' (0.2603), with the correct answer ``e'' ranked third (0.2486). After our latent refinement procedure, the representation is updated, and the model now predicts ``e'' with the highest confidence (0.2626), correcting its earlier mistake. This case shows that even small adjustments in the latent space can significantly shift the model’s final decision.

\subsection{Example 2}

\begin{figure}[htbp]
    \centering
    \includegraphics[width=0.9\linewidth]{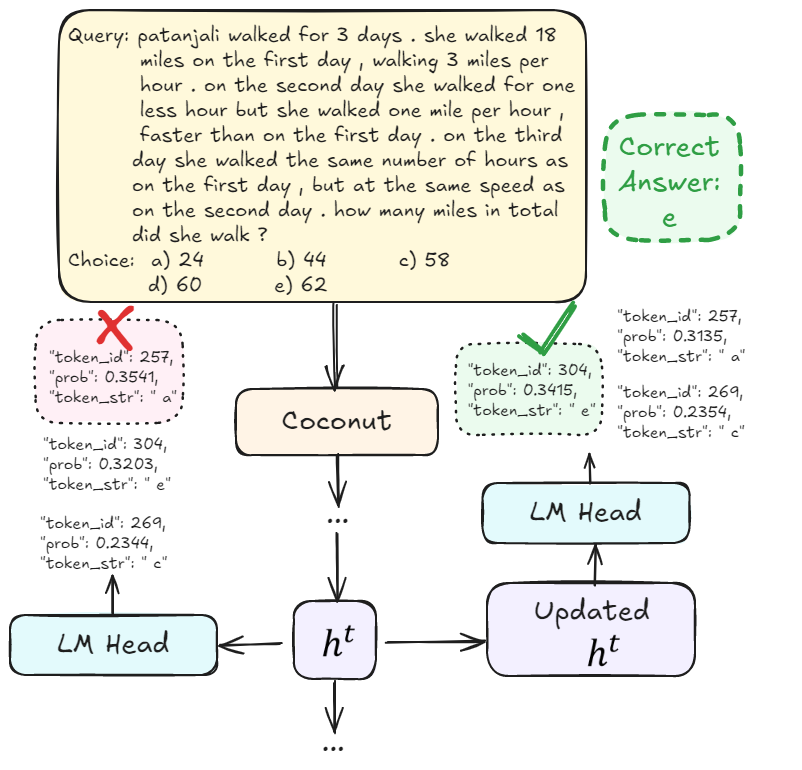}
    \caption{Additional Case study 2: our method adjusts the latent embedding to reach the correct answer.}
    \label{fig:step_case_2}
\end{figure}

\textbf{Figure~\ref{fig:step_case_2}} contains a more complex temporal and numerical reasoning question. Initially, the model incorrectly predicts option ``a'' (0.3541) as the answer. After our latent-space update, the probability of ``e'' rises to 0.3491, making it the top prediction. The model successfully incorporates the revised latent information to override its earlier bias toward an incorrect answer. This example underscores the ability of latent search to shift the model’s internal belief in a non-disruptive and interpretable way.

\subsection{Example 3}

\begin{figure}[htbp]
    \centering
    \includegraphics[width=0.9\linewidth]{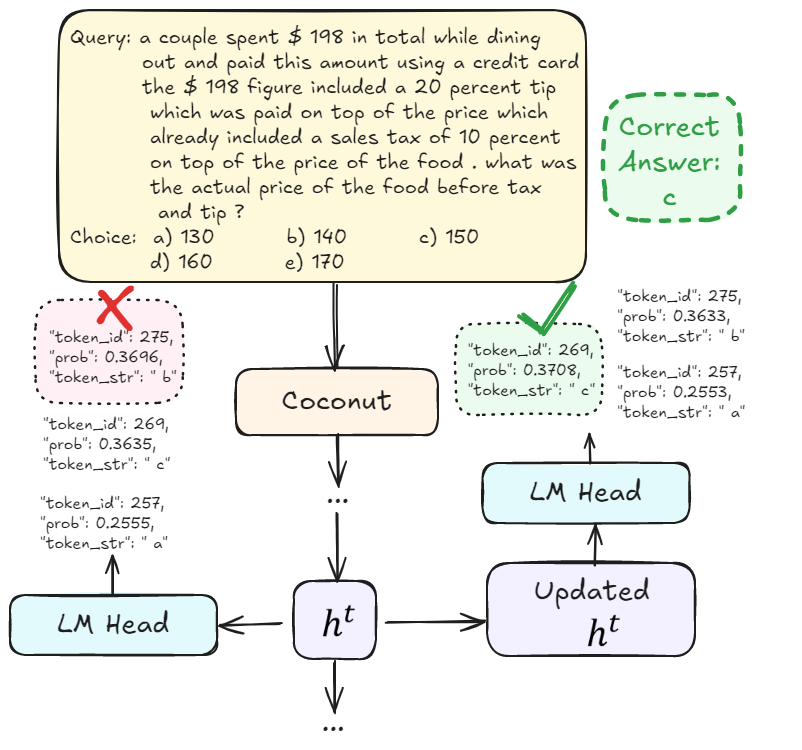}
    \caption{Additional Case study 3: our method adjusts the latent embedding to reach the correct answer.}
    \label{fig:step_case_3}
\end{figure}

In \textbf{Figure~\ref{fig:step_case_3}}, the model is tasked with a multi-step calculation problem involving percentages and sales tax. The initial output ranks ``b'' highest (0.3696), but the correct answer ``c'' is only second (0.3635). After applying contrastive search, the updated latent embedding increases the score of ``c'' to 0.3708, making it the new top choice. Notably, this correction occurs without any retraining or additional supervision.

\section{Limitations}

While our framework improves reasoning performance across multiple tasks and models, it still has several limitations.
(1) Even though the method can be applied to almost all open-source LLMs, we just test it on some relatively small LLMs. So the results are not comparable to the latest model, like ChatGPT-4.1 or DeepSeek-R1. In addition, we do not compare with sampling-based or search-based CoT variants (e.g., Tree-of-Thought~\cite{yao2023tree}, Self-Consistency~\cite{wang2022self}), since our focus is on deterministic, internal latent refinement without additional decoding or re-ranking.
(2) Latent reasoning may not be optimal for all types of tasks. For example, human tend to solve math problems with written-down logic, which involves explicit reasoning processes. In the future study, the routing method~\cite{wang2025mixllm} can be applied to dynamically reason the query.
(3) Latent reasoning lacks human-readable intermediate steps, making it harder to interpret or debug compared to token-based methods.
(4) Inspired by the research of LLM agents~\cite{wang2025dataforge}, future work can introduce tools to ensure stronger, more correct, and more reasonable reasoning processes.

\section{Broader Impact}

The proposed post-training framework to refine the latent reasoning trajectory can also be applied to many downstream applications such as data-centric AI~\cite{wang2025towards,ying2025survey,ying2024feature,ying2024revolutionizing,ying2024unsupervised,ying2023self,ying2024topology,baiprivacy,gong2025agentic,gong2025evolutionary,gong2025sculpting,gong2025neuro,gong2025unsupervised,wang2024knockoff}, linguistics~\cite{ying2020sichuan,wang2022hierarchal}, business~\citep{rs1,wang2024llm,rs2,rs3,li2023sehf}, computer vision \cite{hu2022transrac,dong2023weakly,dong2025mmtok}, and medicine~\cite{liu2019edta,wang2022successful,liu2024calorie,wang2024lcmdc,liu2024pth,li2024sade}.
We can also apply the routing strategy~\cite{wang2025mixllm} to handle situations with multiple reasoning strategies or LLMs. The reinforcement learning strategy~\cite{ying2025bridging} and rule-based augmentation~\cite{bai2025brownian} could also be applied to enhance the robustness of the reasoning process.

\end{document}